          [2001/04/25 1.1 (PWD)]
\documentclass[a4paper,twocolumn]{esapub2005} 
\usepackage{hyperref}
\usepackage{graphicx}
\usepackage{float}
\usepackage{lipsum}  
\usepackage{amsmath}
\usepackage{amssymb}
\usepackage{stmaryrd}
\usepackage{soul}

\usepackage[style=numeric,sorting=none,backend=biber]{biblatex}

\usepackage{xcolor}

\usepackage{times}
\usepackage{url}

\title{Motion Aware ViT-based framework for Monocular 6-DoF Spacecraft Pose Estimation}
\author{Jose Sosa}
\author{Dan Pineau}
\author{Arunkumar Rathinam}
\author{Abdelrahman Shabayek}
\author{Djamila Aouada}
\affil{Interdisciplinary Centre for Security, Reliability and Trust (SnT)}
\affil{University of Luxembourg, Luxembourg}

\begin{document}


\maketitle
\begin{abstract}\vspace{-1em}
Monocular 6-DoF pose estimation plays an important role in multiple spacecraft missions. Most existing pose estimation approaches rely on single images with static keypoint localisation, failing to exploit valuable temporal information inherent to space operations. In this work, we adapt a deep learning framework from human pose estimation to the spacecraft pose estimation domain that integrates motion-aware heatmaps and optical flow to capture motion dynamics. Our approach combines image features from a Vision Transformer (ViT) encoder with motion cues from a pre-trained optical flow model to localise 2D keypoints. Using the estimates, a Perspective-n-Point (PnP) solver recovers 6-DoF poses from known 2D-3D correspondences. We train and evaluate our method on the SPADES-RGB dataset and further assess its generalisation on real and synthetic data from the SPARK-2024 dataset. Overall, our approach demonstrates improved performance over single-image baselines in both 2D keypoint localisation and 6-DoF pose estimation. Furthermore, it shows promising generalisation capabilities when testing on different data distributions.

\end{abstract}
\section{Introduction}\vspace{-1em}

Monocular six-degree-of-freedom (6-DoF) pose estimation is essential for the successful execution of various spacecraft operations, including docking, autonomous rendezvous, in-orbit servicing, and active debris removal~\cite{opromolla2017review}. Hence, achieving accurate and robust 6-DoF pose recovery from monocular images remains a fundamental challenge in computer vision for space applications.

Most existing spacecraft pose estimation (SPE) methods rely on single-image analysis, typically following a two-step pipeline: (1) localising keypoints in a 2D image and (2) estimating the 6-DoF pose using the PnP algorithm~\cite{pauly2023survey}. Unfortunately, this image-by-image approach ignores the rich temporal information available in sequential data. Recent research has shown that incorporating temporal information can improve pose accuracy and enhance robustness~\cite{rondao2022chinet, mohamed2022cubesat, musallam2021leveraging, zuo2024crospace6d, zhang2024monocular}. However, the low diversity of temporal training data in the space domain often constrains the effectiveness of these methods. This scarcity of representative data increases the risk of over-fitting, reducing generalisation to challenging real-world conditions.

\begin{figure}
    \centering
    \includegraphics[width=\linewidth]{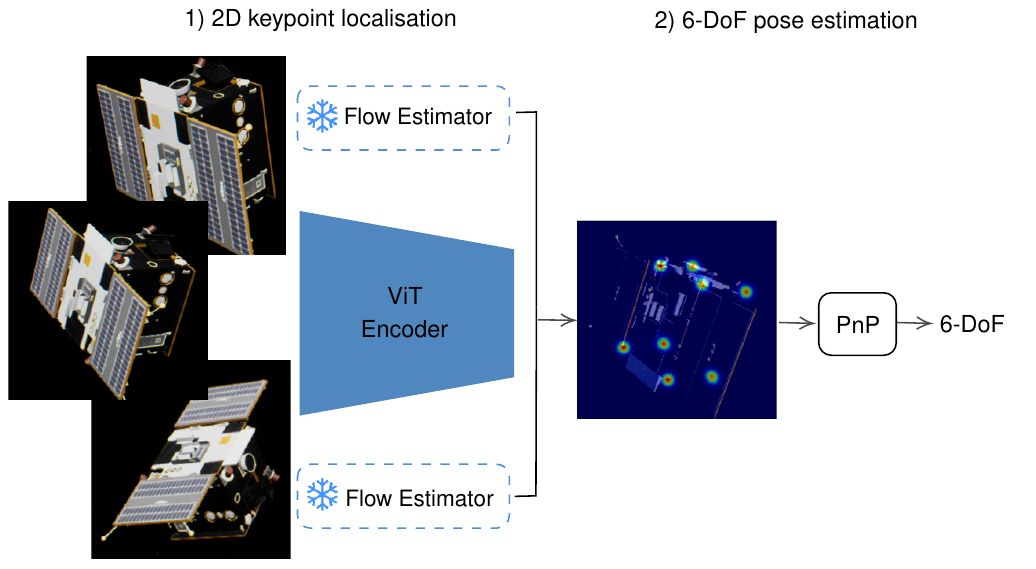}
    \caption{Motion-aware framework for monocular 6-DoF SPE. The model first predicts 2D keypoint locations relying on motion cues from image sequences. Then it uses a PnP solver to recover the full 6-DoF pose.}
    \label{fig:paper-teaser}
\end{figure}

To address these challenges, we propose a motion-aware deep learning framework for monocular 6-DoF SPE from image sequences. Inspired by recent advances in temporal human pose estimation~\cite{song2024motion}, our approach integrates motion-aware heatmaps and a pre-trained optical flow model~\cite{teed2020raft} to capture spacecraft motion. A ViT~\cite{dosovitskiy2020image} serves as the backbone encoder, extracting spatial features. Then, a subsequent module fuses these features with motion information from flow models to predict 2D keypoint locations. During training, the model learns to predict the 2D locations of eight predefined keypoints on the spacecraft structure. After training, PnP recovers the complete 6-DoF pose from the estimated keypoints based on known 2D–3D correspondences.

We train and evaluate our method with the SPADES-RGB dataset, a subset of SPADES~\cite{rathinam2024spades}. This data collection comprises random sequences of spacecraft motion captured in RGB images within a simulated orbital environment. We quantitatively evaluate performance following standard metrics for 2D keypoint localisation and 6-DoF pose accuracy. When using the SPADES-RGB dataset, the proposed approach shows improvements over single-frame baselines in both tasks. Furthermore, experiments on the synthetic and real subsets of the SPARK-2024 dataset~\cite{rathinam_2024_10908215} demonstrate the transfer learning capabilities of our framework across different data distributions. Our main contributions are as follows:

\begin{itemize}\vspace{-1em}
    \item A motion-aware deep learning framework for 6-DoF SPE that combines a ViT encoder with a pre-trained optical flow model to capture temporal correlations in image sequences.\vspace{-0.5em}

    \item A benchmark evaluation of temporal 2D keypoint localisation and 6-DoF pose estimation on the SPADES-RGB dataset.\vspace{-0.5em}  

    \item A cross-domain generalisation analysis, demonstrating the effectiveness of our motion-aware approach when transferring to synthetic and real data on the SPARK-2024 dataset.\vspace{-0.5em}  

\end{itemize}

The remainder of this paper is organised in four sections. Section 2 reviews related work in the field of spacecraft pose estimation. Section 3 details the proposed method, including the keypoint regression network and the pose estimation pipeline. Section 4 presents the experimental results and evaluations on SPADES-RGB and SPARK-2024 datasets. Finally, Section 5 concludes the paper and outlines potential directions for future work.

\section{Related work}\vspace{-1em}

Estimating the 6-DoF pose of objects is a core problem in computer vision, with applications covering robotics manipulation, augmented reality, autonomous driving, and spacecraft navigation~\cite{marullo20236d}. In this work, we focus on 6-DoF SPE from monocular images, a task that is crucial to achieve autonomous on-orbit spacecraft operations. We categorise prior research into two main directions: approaches that rely on single images and those that use multiple images. The latter category is most closely aligned with our work.

\vspace{-1.5em}
\subsection{Single-image spacecraft pose estimation}\vspace{-1em}

In recent years, there has been a significant increase in deep learning-based methods for monocular SPE~\cite{phisannupawong2020vision,huang2021non,proenca2020,black2021real,hu2021wide,li2022learning,wang2022revisiting,lotti2022investigating}. These methods typically fall into two categories: 1) Direct (end-to-end) methods~\cite{phisannupawong2020vision,huang2021non,proenca2020}, which attempt to regress the 6-DoF pose directly from input images. 2) Hybrid methods~\cite{black2021real,hu2021wide,li2022learning,wang2022revisiting,lotti2022investigating}, that decompose the problem into object detection, 2D keypoint regression, and final pose computation using a PnP solver.

Among these, hybrid approaches have emerged as the dominant paradigm due to their increased flexibility, interpretability, and robustness. They typically rely on well-established deep learning backbones as feature extractors. For example, ResNet~\cite{he2016deep}, HRNet~\cite{sun2019deep}, YOLO~\cite{redmon2017yolo9000}, MobileNetV2~\cite{sandler2018mobilenetv2}, and more recently, Transformer-based architectures~\cite{vaswani2017attention,liu2021swin}. Despite their success, most existing methods focus on image-by-image processing, ignoring the temporal continuity inherent in image sequences. This lack of temporal modelling remains a significant limitation, particularly under challenging conditions where robustness and precision are critical~\cite{pauly2023survey}.\vspace{-1em}

\subsection{Multi-image spacecraft pose estimation}\vspace{-1em}
While numerous deep learning approaches estimate spacecraft poses from single images, relatively few learning-based methods explore the use of temporal information for SPE task~\cite{rondao2022chinet, mohamed2022cubesat, musallam2021leveraging, zuo2024crospace6d, zhang2024monocular}.

Recent attempts at temporal SPE are based on single-image pose estimation followed by filtering techniques to track pose over time. For example, methods such as~\cite{park_adaptive_2023, cassinis_leveraging_2023} combine CNN-based keypoint regression with an Unscented Kalman Filter (UKF), introducing adaptive noise models and uncertainty-aware filtering. More recently, Multiplicative Extended Kalman Filters (MEKF) demonstrate greater efficiency and robustness than UKF~\cite{chen_relative_2025, napolano_cnnbased_2025a}. Despite these advances, filtering approaches still rely on single-image keypoint regressions, which lack temporal understanding and might limit accuracy. In other words, these methods only apply temporal constraints on sequences of predicted poses. In addition, filtering techniques are sensitive to initial conditions, the motion model expected in the scenario, and require more observations to converge~\cite{napolano_cnnbased_2025a, chen_relative_2025}.

\begin{figure*}[htbp]
    \centering
    \includegraphics[width=\textwidth]{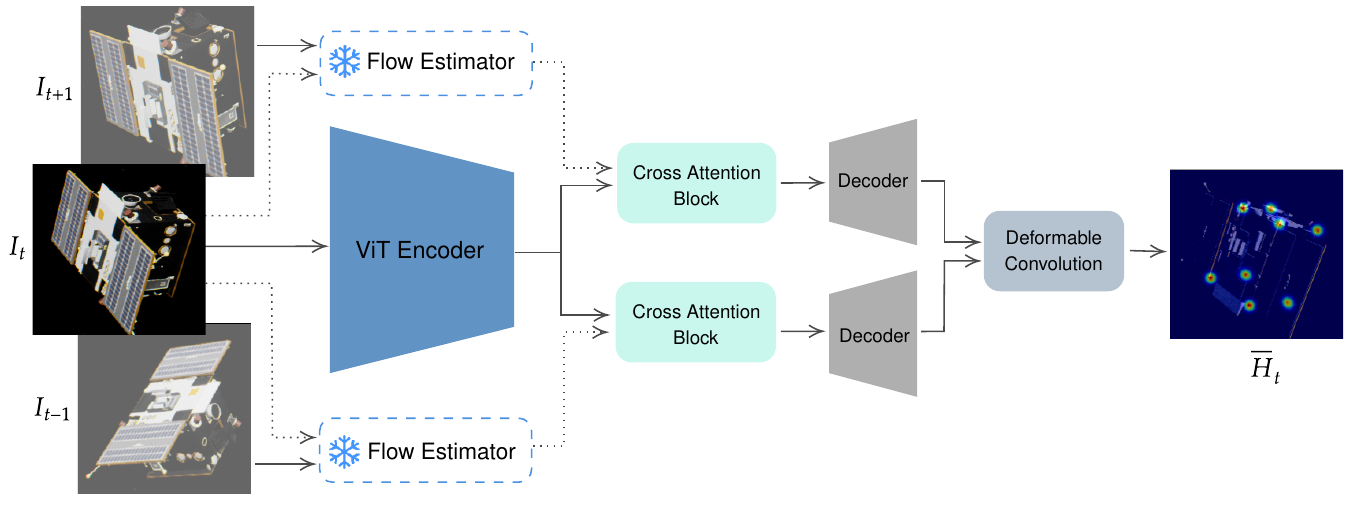}
    \caption{Overview of the proposed method for 2D keypoint localisation. We input three temporally adjacent images $I_{t-1}, I_{t}$, and $I_{t+1}$, a ViT encoder extracts features from the central image, while the other two images support motion vectors generation via a pre-trained flow estimator. Cross-attention blocks fuse these motion vectors and image features, followed by lightweight decoders that produce motion-aware heatmaps. Then, deformable convolutions combine the heatmaps to generate the final pose heatmap $\Bar{H_t}$. Both the final and pairwise motion-aware heatmaps supervise training through a dedicated loss function in line with \cite{song2024motion}.}
    \label{fig:method-diagram}
\end{figure*}

An alternative approach for exploiting temporal data involves using neural networks to directly capture temporal dependencies. For example, ChiNet~\cite{rondao2022chinet} incorporates a CNN-LSTM architecture to regress the 6-DoF pose from image sequences, following an end-to-end strategy. Similarly,~\cite{mohamed2022cubesat} replaces the LSTM with a Temporal Convolutional Network (TCN), which processes temporal features more efficiently in fixed-length sequences~\cite{mishra2017simple}. Unlike end-to-end approaches, Zhang et al.~\cite{zhang2024monocular} utilise a hybrid multi-stage technique for temporal SPE. Initially, they generate supervisory signals for multi-task learning using ground-truth poses. Subsequently, a genetic algorithm is applied to optimise the initial 3D spacecraft model, thereby producing improved pseudo-labels. During inference, temporal consistency is enforced by smoothing the predicted 6-DoF poses. Our proposed approach aligns with this hybrid design but follows a more straightforward pipeline. Instead of applying complex multi-stage refinements on keypoint predictions, we focus on improving them by directly incorporating motion information within the heatmap estimation. This design promotes temporal consistency without additional post-processing.

\vspace{-1em}
\section{Method}\vspace{-1em}
Our proposed approach adopts a hybrid architecture with a two-stage pipeline. It first estimates the 2D locations of keypoints from image features. Then, it predicts the 6-DoF pose from these keypoints given 2D-3D correspondences.

For the first stage, this work builds on the method introduced by~\cite{song2024motion} (initially developed for 2D human pose estimation) and adapt it to the SPE domain. In particular, our proposed approach employs the motion-aware 2D keypoint estimation strategy (MTPose) from~\cite{song2024motion} to incorporate temporal information from consecutive images. In this context, motion information is integrated into keypoint heatmaps, resulting in more accurate keypoint predictions when processing image sequences.\vspace{-1em}

As illustrated by~\autoref{fig:method-diagram}, our approach inputs three temporally adjacent images depicting a moving spacecraft. A ViT encoder extracts features from the target image $I_t$, while the other two $I_{t-1}$ and $I_{t+1}$ serve as temporal references. To capture motion information, the spacecraft’s motion direction is estimated between $I_t$ and its adjacent images using a pre-trained optical flow model~\cite{teed2020raft}. Then, a cross-attention mechanism fuses the motion features with the image features, followed by a pair of lightweight decoders that predict motion-aware heatmaps for each pair of images. Finally, the resulting heatmaps are merged via deformable convolutions to estimate the final heatmap $\Bar{H}_t$ that depicts the 2D keypoint locations in $I_t$.\vspace{-1em}

In the second stage, 2D keypoint locations are extracted from the predicted heatmaps. Then, we feed those to a standard PnP algorithm with RANSAC which computes the spacecraft’s 6-DoF pose in $I_t$. The estimated 6-DoF pose consists of two components denoted as $T = [t_x, t_y , t_z ]$ and $R = [q_w, q_x,q_y, q_z]$ corresponding to translation and rotation, respectively. The following subsections provide more details on the components of our model.\vspace{-1em}


\subsection{2D spacecraft keypoint localisation}\vspace{-1em}
\textbf{Motion-aware heatmaps.} We follow the same motion-aware heatmap generation strategy proposed by~\cite{song2024motion} to incorporate temporal information into 2D keypoint localisation task. In this process each keypoint's heatmap is represented as a 2D elliptical Gaussian distribution, whose shape and orientation encode the direction and magnitude of motion. In particular, the magnitude of motion is computed by measuring the displacements of keypoints between the target frame $I_t$ and its adjacent frames $I_{t-1}$ and $I_{t+1}$. For the motion direction, we compute a rotation angle $\theta$ and use it to rotate the corresponding motion-aware heatmap for each keypoint~\cite{song2024motion}. 

The resulting motion-aware heatmaps $H_{t-1}$ and $H_{t+1}$ encode the direction and magnitude of the motion of each keypoint between the image pairs ($I_t$, $I_{t-1}$) and ($I_t$, $I_{t+1}$), respectively. The standard heatmap $H_t$ simply corresponds to ground truth keypoint locations in the target frame $I_t$. For keypoints that remain static across adjacent images, i.e., below a fixed threshold, we use standard circular Gaussian heatmaps without motion encoding. Due to space constraints, we refer readers to~\cite{song2024motion} for complete implementation details, including the exact formulation of the Gaussian parameters and rotation.

\textbf{Feature extraction.} Our approach relies on a ViT backbone as a feature extractor~\cite{dosovitskiy2020image}. First, the target image $I_t$ is passed through the ViT to obtain features, denoted $F_{I_t}$. For the adjacent images $I_{t-1}$ and $I_{t+1}$, the features are not extracted directly. Instead, these are treated as reference images to assess the spacecraft's motion within the sequence. In this context, the motion between $I_t$ and each supporting image is given by

\begin{equation}
    \begin{aligned}
        V_{t - 1} = O(I_t, I_{t-1})\\  V_{t+1} = O(I_t, I_{t+1})
    \end{aligned}
\end{equation}

where $O$ denotes the pre-trained optical flow estimator from~\cite{teed2020raft}. Since both flow computations involve the central image $I_t$, the notation is simplified by expressing the estimated motion vectors as a function of their second input, i.e., $V_{t-1}$ and $V_{t+1}$. After computing these motion vectors, the corresponding motion-aware features $\bar{F}_{t-1}$ and $\bar{F}_{t+1}$ are generated by combining $V_{t-1}$ and $V_{t+1}$ with the image features $F_{I_t}$, respectively. We follow the MTPose strategy~\cite{song2024motion} for this fusion process. Specifically, we rely on Multi-Head Cross-Attention (MHCA) blocks, where $F_{I_t}$ serves as the query, and each motion vector acts as both key and value. The outputs from MHCA blocks are then passed through a Feedforward Network (FFN) to produce the final motion-aware features $\bar{F}_{t-1}$ and $\bar{F}_{t+1}$.

Given $\bar{F}_{t-1}$ and $\bar{F}_{t+1}$, the corresponding motion-aware heatmaps $\bar{H}_{t-1}$ and $\bar{H}_{t+1}$ are obtained using a decoder $D$, which consists of a couple of deconvolution blocks. Then, those heatmaps are fused via deformable convolutions~\cite{zhu2019deformable} with varying dilation levels. This fusion step produces the final heatmap $\bar{H}_t$, which represents the estimated 2D keypoint positions in the target image $I_t$. To perform supervised training, we rely on a loss function similar to~\cite{song2024motion}. It involves comparing predicted $\bar{H}_t, \bar{H}_{t-1}$ and $\bar{H}_{t+1}$ and ground truth $H_t, H_{t-1}$ and $H_{t+1}$ sets of heatmaps.\vspace{-1em}

\subsection {6-DoF spacecraft pose estimation}\vspace{-1em}
Given the estimated heatmaps $\bar{H}_t$ from the previous stage, we use argmax to extract the corresponding 2D keypoint locations. Then, the 6-DoF pose of the spacecraft is predicted using the PnP algorithm~\cite{lu2018review}. In particular, PnP takes the estimated 2D keypoints, a set of corresponding 3D locations of the keypoints ${X_i}$ in the object reference frame (known from the spacecraft 3D model), and the camera intrinsics matrix $M$. Then, it computes the rotation $R \in \text{SO}(3)$ and translation $T \in \mathbb{R}^3$ that align these points with their corresponding 2D projections ${x_i}$ in the image. Formally, PnP algorithm solves $x_i = M \left(R X_i + T  \right)$, where $i \in [ 1, n ]$ is the index of keypoints. This formulation enables robust 6-DoF pose estimation from monocular images by exploiting both geometric correspondences and prior knowledge of the spacecraft and camera model.\vspace{-1em}

\begin{table*}[htbp]
\centering
\resizebox{0.85\textwidth}{!}{%
\begin{tabular}{l  ccc ccc}
\hline
\textbf{Dataset}  &
\multicolumn{3}{c}{\textbf{2D keypoint localisation metric }} &
\multicolumn{3}{c}{\textbf{6-DoF pose metrics}} \\ 
&  PCK@10 ($\uparrow$) & PCK@5 ($\uparrow$) & PCK@1 ($\uparrow$) & $E_t (\downarrow)$ & $E_R (\downarrow)$ & $E_P (\downarrow)$ \\
\hline
SPADES-RGB &  94.05& 84.05  &18.17 &0.227 &7.57 & 0.157 \\
\hline
SPARK-REAL & 77.45 & 50.93 & 4.24 & 0.210 & 15.31 & 0.320 \\
SPARK-SYN  &  98.16 & 91.45 & 19.17 & 0.137 & 4.69 & 0.099 \\
\hline
\end{tabular}
}%
\caption{2D and 6-DoF pose estimation metrics. Quantitative results from our model on SPADES-RGB, SPARK-REAL, and SPARK-SYN evaluation sets.}
\label{table:main-results}
\end{table*}

\section{Experiments}\vspace{-1em}
\subsection{Data}\vspace{-1em}
\textbf{SPADES-RGB.} 
We utilise SPADES dataset~\cite{rathinam2024spades} for both training and evaluation. This dataset features simulated data utilising the Proba-2 satellite from the PROBA-2 mission~\cite{gantois2006proba}, collected in the Zero-G lab~\cite{olivaresmendez_zerog_2023} with a scaled mock-up model. SPADES offers sequences of temporal images depicting diverse background and lighting conditions. Although, originally designed for event-based vision research, the dataset also includes RGB imagery. In this paper, we exclusively employ the RGB part, hence referred to as SPADES-RGB. The dataset contains 300 sequences and each sequence consists of approximately 600 RGB images. These sequences comprise training, validation, and test sets, with 210, 45, and 45 sequences assigned to each group, respectively.

\textbf{SPARK-2024 Stream 2.} To assess the generalisation of our approach, we extend the evaluation to the SPARK-2024 dataset~\cite{rathinam_2024_10908215}. Similar as SPADES, this dataset features the same Proba-2 satellite, but it includes two different domains, real and synthetic. The real data comprises four sequences with a total of 2,048 images. In contrast, the synthetic data is more extensive, including 7,424 images distributed across 99 sequences. Note that the SPARK-2024 dataset has been solely used for evaluation purposes and not for training the models.

\vspace{-1em}
\subsection{Metrics}\vspace{-1em}
To quantitatively evaluate the performance of the model, we evaluate two tasks, 2D keypoint localisation and 6DoF pose estimation. For 2D keypoint localisation, we use the normalised Percentage of Correct Keypoints (PCK) metric. Given the estimated 2D keypoints and the corresponding ground truth keypoints, PCK measures the proportion of keypoints that fall within a specified distance threshold, normalised by the size of the bounding box. Following standard practices on 2D keypoint localisation, the normalisation factor is defined as the diagonal length of the bounding box enclosing the predicted keypoints. The PCK metric is computed at three different thresholds, including 1\%, 5\%, and 10\% of this diagonal, denoted as PCK@1, PCK@5, and PCK@10, respectively.

For evaluating 6-DoF pose estimation, we rely on standard pose error metrics~\cite{park2022speed+}, including position error $E_t$, orientation error $E_q$, and pose error $E_{P}$. The following equations formally define these metrics,

\begin{equation}
    \begin{aligned}
        E_t &= || \hat{T} - T ||_2, \\
        E_q &= 2 \ \text{arccos} \ | <\hat{R}, R> |, \\
        E_{P} &= E_q + E_t / ||T||,
    \end{aligned}
\end{equation}

where $T$ and $\hat{T}$ represent the ground truth and estimated positions, respectively, and $R$ and $\hat{R}$ denote the ground truth and predicted quaternions.\vspace{-1em}

\subsection{Implementation details}\vspace{-1em}
During training, our model inputs three consecutive images, each derived from a cropped section of the original image.
The cropped region is centred on the spacecraft's bounding box, which we assume is provided as part of the dataset annotations.
Within each dataset sequence, the frames are sampled with a seven-frame interval to encourage the model to capture meaningful motion dynamics. We rely on a ViT-Base architecture as encoder, initialised with ImageNet weights~\cite{deng2009imagenet}. Unlike the approach in~\cite{song2024motion}, the encoder is fine-tuned jointly with the other components of our model, except for the optical flow estimator. For flow estimation, we adopt a pre-trained RAFT-Large model~\cite{teed2020raft}, which is kept frozen throughout the training process. We utilise Adam optimiser with a learning rate of $1.5 \times 10^{-4}$, a cosine learning rate scheduler, and a batch size of 128. Training runs for 150 epochs on 4 NVIDIA A100-40GB GPUs. For 6-DoF pose estimation, we follow the PnP implementation provided in OpenCV\footnote{\url{https://docs.opencv.org/4.x/d5/d1f/calib3d_solvePnP.html}}. All other implementation details, including the loss function and motion-aware heatmap generation, follow the setup in \cite{song2024motion}.\vspace{-1em}

\begin{figure}[ht]
    \centering
    \includegraphics[width=\linewidth]{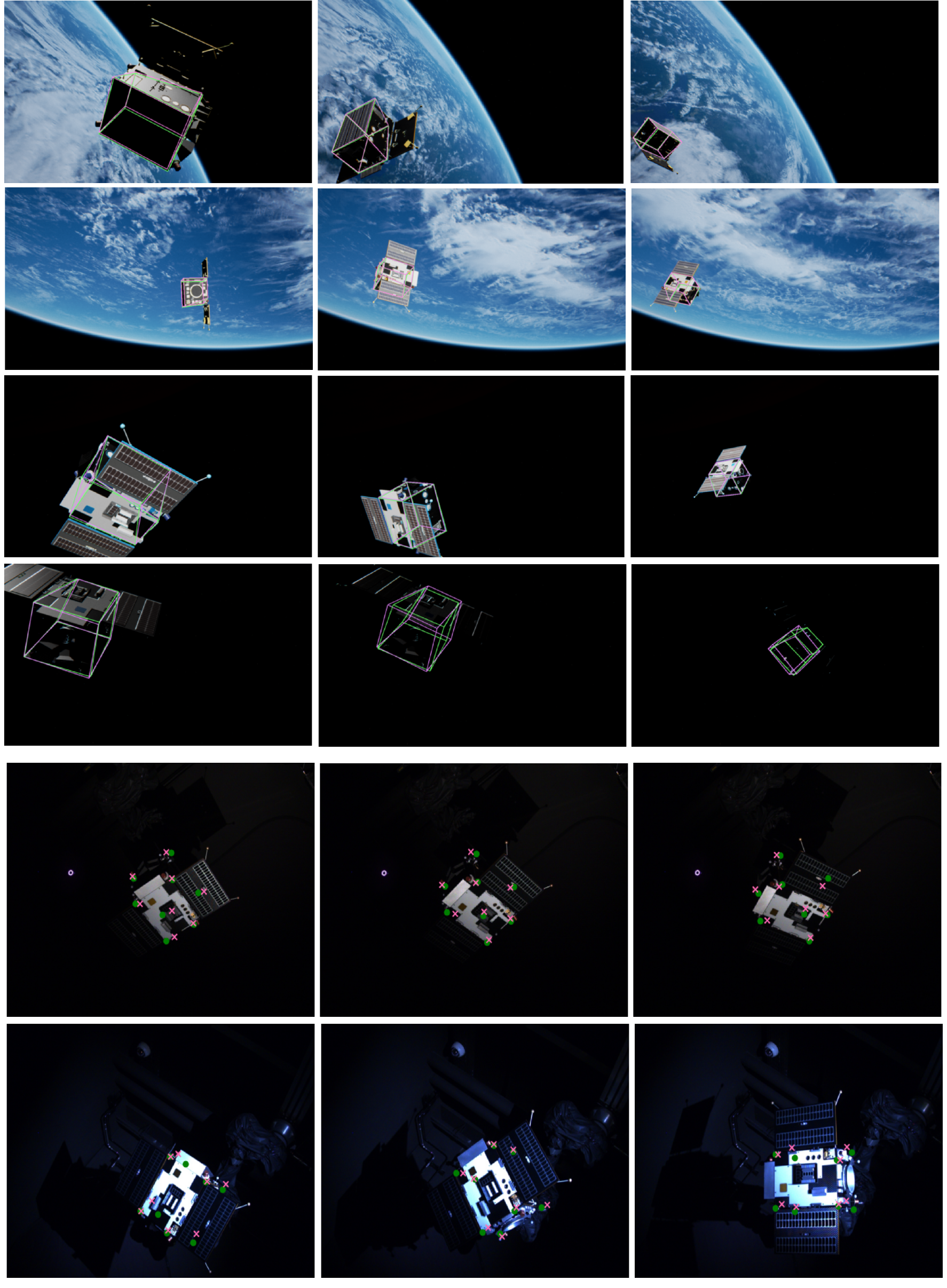}
    \caption{Qualitative results for 2D keypoint localisation and 6-DoF SPE. The first four rows depict sequences from SPADES-RGB data with their corresponding 6-DoF ground truth (green) and estimated poses (pink). Last two rows show real sequences from SPARK-2024 depicting ground truth (green) and estimated (pink) 2D keypoint locations. Zoom in for better visualisation.}
    \label{fig:vis}
\end{figure}

\subsection{Results and discussions}\vspace{-1em}

We evaluate the trained model on both keypoint localisation and 6-DoF pose estimation, using the 45 test sequences from the SPADES-RGB dataset. ~\autoref{table:main-results} reports the PCK metrics at several thresholds and the 6-DoF pose metrics. For the 2D keypoint localisation task, note that reducing the threshold as in PCK@5 does not significantly impact overall performance. However, when using PCK@1, the performance drops noticeably, as this stricter threshold poses a challenge for most pose estimators. Since previous works have only reported 6-DoF results on the event-based version of the SPADES dataset~\cite{rathinam2024spades}, our results on the 6-DoF pose estimation task provide a benchmark for the image-based version, SPADES-RGB.

\begin{figure}[ht!]
    \centering
    \includegraphics[width=\linewidth]{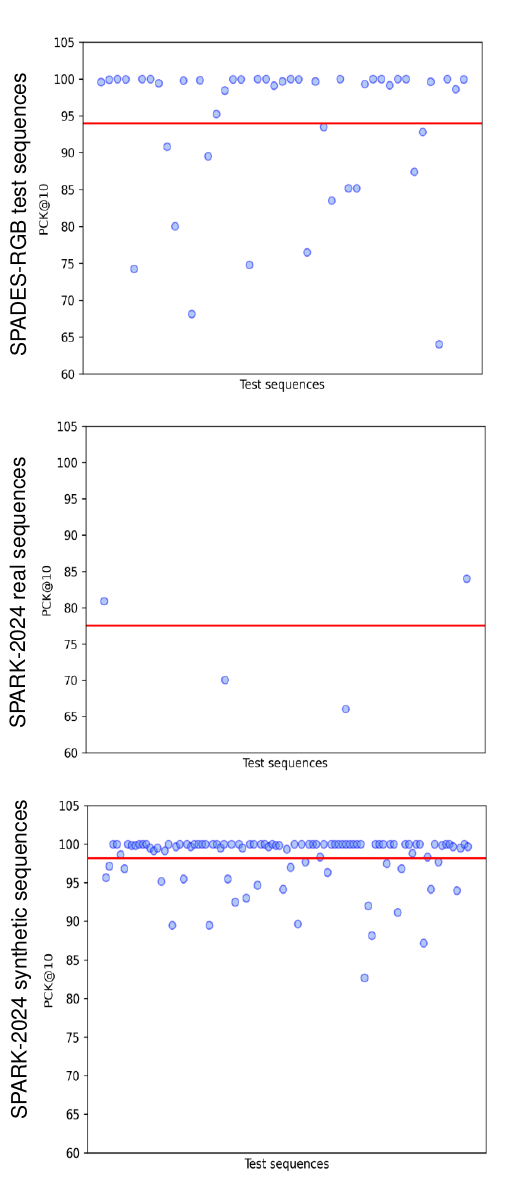}
    \caption{Distribution of PCK@10 scores per each sequence on different test data. Each blue circle in the plots represents a sequence from the corresponding set, namely SPADES-RGB, SPARK-REAL and SPARK-SYN. Red lines correspond to the average PCK@10 for each set. }
    \label{fig:sequences-dist}
\end{figure}

To further assess the generalisation capabilities of our approach, we test our trained model, without any fine-tuning or domain adaptation, on both the synthetic and real subsets of the SPARK-2024 dataset. As can be observed in~\autoref{table:main-results}, metrics remain relatively consistent across SPADES-RGB and SPARK-SYN. However, performance degrades on SPARK-REAL, with all metrics showing a decline. This drop on performance is expected, due to the data distribution shift between synthetic and real sets. ~\autoref{fig:sequences-dist} provides per-sequence analysis of 2D keypoint localisation.

Interestingly, the model performs better on the SPARK-2024 synthetic data than on SPADES-RGB. This might be due to the fact that both datasets use the same spacecraft model. Futhermore, the synthetic sequences from SPARK-2024 likely present fewer challenges in terms of pose variation and illumination. In contrast, the SPADES-RGB dataset includes more difficult scenarios, as further illustrated by the qualitative examples in~\autoref{fig:vis}.




\vspace{-1em}

\subsection{Ablations}\vspace{-1em}
We perform an ablation study on two high-level components of our approach, as depicted in~\autoref{table:ablation}. The first experiment involves training a version of the model using only a single input image (N-MA), effectively removing motion awareness. As shown in the second row of~\autoref{table:ablation}, this modification leads to a consistent drop in performance, particularly under stricter PCK thresholds. The 6-DoF pose estimation is also affected, showing higher error values compared to the full model that uses three input images. This experiment showcases the positive effect of relying on temporal information for this approach.

\begin{table}[ht!]
\centering
\resizebox{\columnwidth}{!}{%
\begin{tabular}{lccccc}
\hline
\textbf{Setup} & PCK@10 $\uparrow$ &PCK@5$\uparrow$ & $E_t \downarrow$ & $E_R \downarrow$ & $E_P \downarrow$ \\
\hline
 MA &  \textbf{94.05}& \textbf{84.05} & \textbf{0.227} &\textbf{7.57} &\textbf{0.157}\\
 \hline
 N-MA & 92.63 & 70.33  & 0.375 & 8.84 & 0.197 \\
 FE & 15.76  & 8.37 & 4.88 & 104.42 & 2.37 \\
\hline
\end{tabular}%
}
\caption{Ablation study of model components. 2D and 6-DoF pose estimation metrics under two training settings: a baseline without motion awareness (N-MA), and training the motion-aware framework with a frozen ViT encoder (FE).}
\label{table:ablation}
\end{table}

Additionally, we evaluate a variant of the model where the encoder is kept frozen and only the remaining components are trained. This configuration (FE) results in very poor performance as depicted in the second row of~\autoref{table:ablation}. We hypothesise that ImageNet-pretrained weights fail to capture the domain-specific features of the spacecraft, which are crucial for our task. Note that in the original model for human pose estimation by Song et al.~\cite{song2022deep}, the authors also freeze the encoder. However, they initialise the ViT encoder from a pre-trained human pose estimation model, not from ImageNet.

We conduct an additional analysis to breakdown the keypoint localisation performance. We simply filter the predictions based on their PCK@10 score and evaluate the proportion of the data that meet that condition. As illustrated in \autoref{table:pck-filtering}, the PCK and 6-DoF metric is computed only on the subset of samples that satisfies the condition of PCK@10 scores. Surprisingly, around 81\% of our predictions achieve a PCK@10 greater than 90\%, indicating that, in most cases, the model accurately estimates the majority of keypoints. 

\begin{table}[ht!]
\centering
\resizebox{\columnwidth}{!}{%
\begin{tabular}{lccccc}
\hline
\textbf{Setup} & Data (\%) & PCK@10 $\uparrow$ & \textbf{$E_t \downarrow$} & \textbf{$E_q \downarrow$} & \textbf{$E_{P} \downarrow$} \\
\hline
No filtering & 100 & 94.05 & 0.2275 & 7.57 & 0.1571 \\
\hline
PCK$>$12.5   & 99.04 & 94.91 & 0.2186 & 6.85 & 0.1437 \\
PCK$>$25     & 98.18 & 95.51 & 0.2105 & 6.53 & 0.1375 \\
PCK$>$50     & 95.36 & 97.01 & 0.1811 & 5.75 & 0.1209 \\
PCK$>$90     & 81.50 & 100.00 & 0.1343 & 4.63 & 0.0969 \\
\hline
\end{tabular}%
}
\caption{Effect of PCK-based filtering on pose metrics. PCK@10 and 6-DoF pose accuracy after applying different filters to the predictions.}
\label{table:pck-filtering}
\end{table}

\vspace{-1em}
\section{Conclusion}\vspace{-1em}

This work presents a framework for estimating 6-DoF spacecraft poses from monocular image sequences. Our approach follows a hybrid design. It first estimates 2D keypoints while directly incorporating motion information from supporting images. Then, it computes the 6-DoF pose using a PnP solver applied to the predicted keypoints. We demonstrate the effectiveness of our method and set a benchmark on the SPADES-RGB dataset. Additionally, the generalisation performance of the model is shown through cross-dataset testing on both real and synthetic data from the stream 2 of SPARK-2024 dataset. The results indicate robust performance in both 2D keypoint localisation and 6-DoF pose estimation across domains. Future work contemplates to adapt the method to longer image sequences, which might help to capture richer motion dynamics and potentially improve pose estimation performance.

\vspace{-0.5em}
\textbf{Acknowledgments.} Authors would like to thank Samet Hicsonmez, Nidhal Eddine Chenni, Inder Pal Singh, and Mohammed Guermal for great discussions. This research has been conducted in the context of the DIOSSA project, supported by the European Space Agency (ESA) under contract no. 4000144941241NL/KK/adu. Experiments were performed on Luxembourg national supercomputer MeluXina. Thanks to LuxProvide teams for their support.

{
    \printbibliography
}

\end{document}